\documentclass[letterpaper]{article} % DO NOT CHANGE THIS
\usepackage{aaai24}  % DO NOT CHANGE THIS
\usepackage{times}  % DO NOT CHANGE THIS
\usepackage{helvet}  % DO NOT CHANGE THIS
\usepackage{courier}  % DO NOT CHANGE THIS
\usepackage[hyphens]{url}  % DO NOT CHANGE THIS
\usepackage{graphicx} % DO NOT CHANGE THIS
\usepackage{natbib}  % DO NOT CHANGE THIS AND DO NOT ADD ANY OPTIONS TO IT
\usepackage{caption} % DO NOT CHANGE THIS AND DO NOT ADD ANY OPTIONS TO IT
\usepackage{algorithm}
\usepackage{algorithmic}

\usepackage{newfloat}
\usepackage{listings}

\usepackage{color}
\usepackage{amsfonts,amssymb,amsmath} 
\usepackage{xspace}
\usepackage{booktabs}
\usepackage{multirow}
\newcommand{\themodel}{HeterFC\xspace}
\DeclareCaptionStyle{ruled}{labelfont=normalfont,labelsep=colon,strut=off} % DO NOT CHANGE THIS
\floatstyle{ruled}
\newfloat{listing}{tb}{lst}{}
\floatname{listing}{Listing}
\title{Heterogeneous Graph Reasoning for Fact Checking over Texts and Tables}
\author {
    Haisong Gong\textsuperscript{\rm 1,\rm 2},
    Weizhi Xu\textsuperscript{\rm 3},
    Shu Wu\textsuperscript{\rm 1,\rm 2}\thanks{Corresponding Author},
    Qiang Liu\textsuperscript{\rm 1,\rm 2},
    Liang Wang\textsuperscript{\rm 1,\rm 2}
}
\affiliations {
    \textsuperscript{\rm 1}Center for Research on Intelligent Perception and Computing\\
State Key Laboratory of Multimodal Artificial Intelligence Systems\\
Institute of Automation, Chinese Academy of Sciences\\
    \textsuperscript{\rm 2}School of Artificial Intelligence, University of Chinese Academy of Sciences\\
    \textsuperscript{\rm 3}ByteDance Inc.\\
    gonghaisong2021@ia.ac.cn, jason\_xu1998@163.com, \{shu.wu, qiang.liu, wangliang\}@nlpr.ia.ac.cn
}

\begin{document}

\maketitle

\begin{abstract}
Fact checking aims to predict claim veracity by reasoning over multiple evidence pieces. It usually involves evidence retrieval and veracity reasoning. In this paper, we focus on the latter, reasoning over unstructured text and structured table information. Previous works have primarily relied on fine-tuning pretrained language models or training homogeneous-graph-based models. Despite their effectiveness, we argue that they fail to explore the rich semantic information underlying the evidence with different structures. To address this, we propose a novel word-level Heterogeneous-graph-based model for Fact Checking over unstructured and structured information, namely \themodel. Our approach leverages a heterogeneous evidence graph, with words as nodes and thoughtfully designed edges representing different evidence properties. We perform information propagation via a relational graph neural network, facilitating interactions between claims and evidence. An attention-based method is utilized to integrate information, combined with a language model for generating predictions. We introduce a multitask loss function to account for potential inaccuracies in evidence retrieval. Comprehensive experiments on the large fact checking dataset FEVEROUS demonstrate the effectiveness of \themodel. Code will be released at: https://github.com/Deno-V/HeterFC.
\end{abstract}

\section{Introduction}\label{sec:introduction}
Fact checking, or fact verification, predicts claim veracity using evidence. This task has practical applications in various domains like politics \cite{liu2023out, Xu2022EvidenceawareFN}, news media \cite{zellers2019defending}, public health \cite{Naeem2020TheC, Krause2020FactcheckingAR}, and science \cite{Wright2022GeneratingSC,wadden2020fact}, attracting extensive research. Prior efforts predominantly address unstructured fact checking, handling evidence and claims as plain text \cite{thorne2018fever, wang2017liar,xu2023counterfactual}. However, real-world contexts often involve structured data like tables, creating a pressing need for fact checking across unstructured and structured information.

Fact checking mainly involves evidence retrieval and veracity reasoning, which are two independent tasks \cite{aly2021feverous}. The aim of evidence retrieval is to retrieve as much the claim-related evidence (golden evidence) as possible; the target of veracity reasoning is to precisely predict the veracity of claim based on the retrieved evidence from the former stage. In this paper, we mainly focus on the design of veracity reasoning model.

Previous veracity reasoning models \cite{aly2021feverous, kotonya2021graph, funkquist2021combining,hu2022dual,bouziane2021fabulous,zhao2020transformer,wu2022adversarial} have primarily relied on fine-tuning pretrained language models or training homogeneous-graph-based models. In the fine-tuning approach, they firstly transform tables into sentences via some heuristic linearizing rules. Then, a pretrained language model (PLM), such as RoBERTa \cite{liu2019roberta}, is fine-tuned by concatenating all pieces of evidence as the input. In the homogeneous-graph-based approach, they construct a homogeneous fully-connected evidence graph where each node is treated as a piece of evidence. After that, a graph neural network (GNN) is utilized to propagate neighborhood information, which enables the semantic representations of different pieces of evidence to be aggregated.

While effective, existing approaches exhibit two key weaknesses. \textit{Firstly}, fact checking necessitates capturing semantics among various evidence pieces, demanding intricate modeling of evidence relationships. Transformer-based methods often fall short as they merely concatenate evidence or deal with point-wise claim-evidence pairs, insufficiently exploring complex evidence interconnections. \textit{Secondly}, prevalent graph-based methods construct sentence-level graphs with claim-evidence pairs as nodes, employing Pre-trained Language Models (PLMs) for node representations \cite{zhou2019gear, liu2020fine, kotonya2021graph}. Although these models perform well in conventional fact checking, they falter in scenarios involving both structured and unstructured information. This is due to the limitations of sentence-level graphs in capturing fine-grained details such as entities and time phrases. Furthermore, assuming uniform relationships between node pairs overlooks the diverse properties inherent in table and sentence evidence.

To tackle the aforementioned problems, we propose a novel word-level \underline{\textbf{Heter}}ogeneous-graph-based model for \underline{\textbf{F}}act \underline{\textbf{C}}hecking over unstructured and structured information, \themodel for brevity. Specifically, we first construct a graph where nodes represent words in all pieces of evidence, thereby achieving a granularity at word-level. Then, to capture the different relationships in structured and unstructured information, we specially design three kinds of connections on the graph, namely intra-sentence edges, intra-table edges, and inter-evidence edges. 
In detail, intra-sentence edges and intra-table edges are added between each word and its local contextual words in a fix-sized sliding window. Inter-evidence edges are between the same keyword appearing in several pieces of evidence, which allows the important information be aggregated across the evidence via these edges.
We employ the relational graph convolutional network (R-GCN) to perform neighborhood propagation and readout the representations of each evidence, followed by an attention mechanism to obtain information from all pieces of evidence. Combined with a language model, we get the final veracity prediction. To train the model, in addition to cross-entropy loss for the claim veracity, we propose a multitask loss to assist the model in discerning between valid and invalid evidence, thereby enhancing the overall performance of the model.

In a nutshell, our main contributions can be listed as follows,
\begin{itemize}
    \item We figure out the inapplicability of previous homogeneous-graph-based methods in the traditional unstructured fact checking and analyze the underlying possible reasons.
    \item We propose a novel word-level heterogeneous-graph-based model, namely \themodel, which is specially designed for fact checking over unstructured and structured information.
    \item Extensive experiments on the large-scale FEVEROUS fact-checking dataset, which includes both structured and unstructured information, have demonstrated the effectiveness of our proposed model over several baselines.
\end{itemize}

\section{Related Work}

\subsection{Fact Checking Over Unstructured Information}
Fact checking, a form of natural language inference (NLI), involves predicting claim veracity by reasoning over multiple evidence pieces. Existing methods fall into two categories. The first category uses pretrained language models (PLMs), fine-tuning them for fact checking.  They organize the input by concatenating all evidence and claim into a single sentence \cite{aly2021feverous}, or processing each evidence separately with the claim using aggregation techniques \cite{soleimani2020bert,gi2021verdict}. 
The second category employs graph neural networks (GNNs) to capture complex semantic interactions. GEAR \cite{zhou2019gear} constructs sentence-level fully-connected evidence graphs with GNNs. \citet{zhao2020transformer} use Transformer-XH for graph representation, while \citet{liu2020fine} introduce KGAT with node and edge kernels. DREAM \cite{zhong2020reasoning} incorporates semantic role labeling for fine-grained semantic graphs. \citet{chen2022evidencenet} propose EvidenceNet with symmetrical interaction attention and gating on sentence-level evidence graphs.

\subsection{Fact Checking Over Mixed-type Information}
Unlike unstructured fact checking, fact checking over both structured and unstructured information requires handling a combination of structured tables and unstructured text. 
Existing methods include table linearization, where tables are converted into text, potentially losing structural information \cite{gi2021verdict,kotonya2021graph, malon2021team}. Another approach is to combine sentence and table evidence using models like TAPAS \cite{funkquist2021combining,bouziane2021fabulous,hu2022dual}. In graph-based methods, previous works focused on homogeneous sentence-level evidence graphs \cite{kotonya2021graph}. In contrast, our approach introduces word-level nodes and heterogeneous relations, making it more suitable for fact checking over structured and unstructured information.

\subsection{Heterogeneous Graph Neural Networks}
Heterogeneous graph neural networks (heterGNNs) are specialized GNNs designed for neighborhood propagation on graphs with different types of edges. R-GCN \cite{schlichtkrull2018modeling} is a representative heterGNN that assigns trainable weight matrices to each relation. HeterGNNs have been successfully applied in various domains including recommender systems \cite{fan2019metapath, zhao2017meta, yan2021relation} and question answering \cite{yu2019heterogeneous, sun2018open}. 

\section{Method}
\begin{figure*}[t]
\setlength{\abovecaptionskip}{0pt}  % reduce gap
\setlength{\belowcaptionskip}{0pt}  % reduce gap
   \begin{center}
%   \vspace{-4mm}
   \includegraphics[width=1\textwidth]{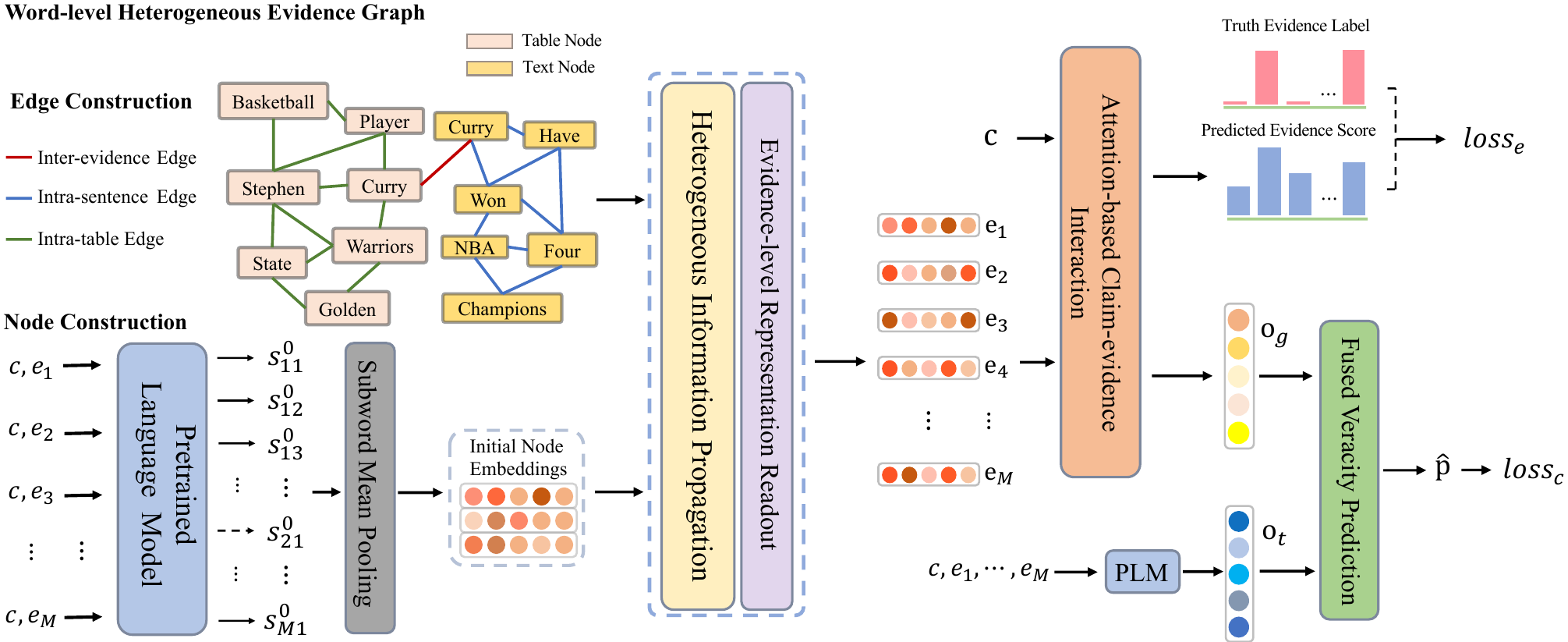}
   \end{center}
   % \vspace{0cm}
   \caption{Architecture of \themodel. Inputs are claim \(c\) and an evidence set \(\mathcal{E}\). There are five main parts: 1) \emph{Word-level evidence graph construction.} Initial node embeddings are obtained by using PLM and subword mean pooling. Three types of edges are designed for the heterogeneous connection. 2) \emph{Heterogeneous information propagation.}  R-GCN is used to perform neighborhood aggregation on the word-level evidence graph. 3) \emph{Evidence-level representation readout.}  Evidence representations are obtained by pooling over subgraphs corresponding to each piece of evidence. 4) \emph{Attention-based claim-evidence interaction.}  Graph representation \(\mathbf{o}_g\) is generated by claim-guided evidence combination. A supervised loss item, \(Loss_{e}\), is computed based on the attention assignment.
   5) \emph{Fused veracity prediction.} The claim and evidence are concatenated and fed into PLM to obtain \(\mathbf{o}_t\), which, when combined with the graph representation \(\mathbf{o}_g\), forms the final representation. A fully-connected network takes the representation as input and generates the prediction \(\mathbf{\hat p}\).  \themodel is trained using classification loss \(Loss_c\) and assisted \(Loss_e\). } 
   \label{fig:model}
\end{figure*}

In this section, we introduce the proposed method \themodel in details. The overall framework is shown in Figure \ref{fig:model}.

\subsection{Task Formulation}
The aim of multi-structured fact checking is to predict the veracity of a claim according to several pieces of evidence, which contains both tables and texts.
Mathematically, given a claim \(c\) and a retrieved evidence set \(\mathcal{E} = \{e_1, e_2, \ldots, e_M\}\), where each piece of evidence \(e_i\) represents either a sentence or a table cell, we need to propose a model \(\mathbf{\hat p} = f(c, \mathcal{E})\) to output the predicted veracity \(\mathbf{\hat p}\).
\subsection{Word-level Evidence Graph Construction}
In this part, we elaborate the design of nodes and edges on the evidence graph.
\subsubsection{Node Construction}
We treat each word in the evidence as a node on the evidence graph, since it contains more fine-grained semantic information than the sentence.\footnote{We have tried token-level node construction, which is more fine-grained than the word-level nodes. However, it is less effective and more analysis can be seen in the experimental section.} 
To achieve this, we employ a PLM to process each claim-evidence pair to build the initial node representations. 

The sentence evidence can be directly treated as input to a PLM, however, since tables have a distinct structure from sentences, table evidence can't be directly processed by PLM. To address this, we adopt the idea of cell linearization proposed in \cite{hu2022dual}. Specifically, each piece of table evidence is transformed into either ``\textit{\textless column header\textgreater\  for \textless row header\textgreater\  is \textless cell value\textgreater\  }'' if it belongs to a general table or ``\textit{\textless column header\textgreater\  :\textless row header\textgreater\  of \textless title\textgreater\  is \textless cell value\textgreater\ } '' if it is from an infobox-type table. 
%For example, the table in Figure \ref{fig:example} is an infobox-type table, the cell ``\textit{Golden State Warriors}'' should be transformed into ``\textit{Carrer information: Team of Stephen Curry is Golden State Warriors }''.

By applying cell linearization technique to table evidence, we are able to feed each claim-evidence pair to a PLM and obtain the embedding of each token (subword) in both table evidence and sentence evidence. It is noteworthy that though claim-evidence pair is fed into PLM, the subwords in claim are excluded and only the subwords in evidence are kept. This can be expressed mathematically as follows:
\begin{align}
    \mathbf{S}^0_i &= \text{PLM}([c, e_i]) \\
    \mathbf{S}^0_i &:= \mathbf{s}^0_{i0} \ || \ \mathbf{s}^0_{i1} \ || \ \ldots \ || \ \mathbf{s}^0_{ij} 
\end{align}
where PLM is a RoBERTa model here, following previous works \cite{aly2021feverous}. \(c\) denotes the claim and \(e_i\) is the \(i\)-th evidence. \(\mathbf{s}^0_{ij}\) represents the embedding of the \(j\)-th subword in the \(i\)-th evidence and \(||\) is the concatenation of vectors.

Then, we generate the embedding of a whole word via computing the mean of the embeddings of its corresponding subwords. By following this approach, we can obtain the word embedding matrix \(\mathbf{H}^0\) for the all pieces of evidence by the mentioned way, where \(\mathbf{H}^0\) is the initial node representations since we treat each word in the evidence as a node.

\subsubsection{Edge Construction}
After constructing word-level nodes, the next step is to design the connections among them. The simplest way is to construct a fully-connected graph, where each node shares an edge with every other node on the graph. However, this approach may bring too much noise since only part of information is related and beneficial for a node.
Especially, the input evidence inevitably contains some unrelated information due to the error of the retrieval model. Therefore, a more elaborate design is required in this task.

We carefully design three types of edges to capture the heterogeneous information among several pieces of evidence. Specifically, the three types of edges are named inter-evidence edges, intra-sentence edges and intra-table edges. The illustration of such edges is shown in Figure \ref{fig:model} where different type of edge is illustrated with different color in the graph, and we introduce them in detail as follows,

\textbf{Inter-evidence edges \(r_e\). \ }
Aggregating relevant information from multiple pieces of evidence is crucial for accurate claim veracity prediction. In fact verification scenarios involving both texts and tables, it is common for the same entity to be referenced in different types of evidence. Thus, integrating information from both sources is necessary for a comprehensive understanding. To capture the multi-hop relationship between evidence, we construct edges connecting the same word in different pieces of evidence. By propagating information along these edges, we can capture the flow of information between related evidence. To ensure the quality of inter-evidence edges, we filter out stop words like ``is,'' ``of,'' and ``the'' to prevent constructing edges between frequently used but insignificant words.

\textbf{Intra-sentence edges \(r_s\). \ } A word in the sentence is usually associated with its local context for understanding the semantics \cite{mikolov2013distributed}. Therefore, we adopt this traditional technique and employ a sliding window with a fix size \(w\) to cover the local context. In this way,
each word in the center of the window has edges with the rest of words in the window, through which each word is connected with its context on the graph. Thus, the contextual information can be aggregated via a one-layer GNN, which is beneficial for learning the sentence-level semantics.   

\textbf{Intra-table edges \(r_t\). \ } 
Tables have a completely different structure compared with sentences. The header and the cell form a key-value relationship. This structure is distinct from that of a sentence, which contains many stop words (is, the, etc.) to ensure fluency. 
Based on the analysis above, we assign edges among the cell, its row header, column header and its page title. To achieve this, we reuse the cell linearization method mentioned in the node construction section to transform each table cell into sequence and utilize the fix-sized window again to connect words, just like building intra-sentence edges.

Eventually, we construct a word-level evidence graph \(\mathcal{G}\) via the aforementioned design, which involves three different relations \(\mathcal{R} = \{r_s, r_t, r_e\}\). Next, we introduce the main model architecture.

\subsection{Heterogeneous Information Propagation}
The constructed evidence graph includes various edges, making homogeneous GNNs unsuitable. Thus, we employ relational graph convolutional networks (R-GCN) to capture distinct node relations. Formally, it can be written as, 
\begin{equation}
\mathbf{h}_{i}^{l+1}=\sigma\left(\sum_{r \in \mathcal{R}} \sum_{j \in \mathcal{N}_{i}^{r}} \frac{1}{c_{i, r}} \mathbf{W}_{r}^{l} \mathbf{h}_{j}^{l}+\mathbf{W}_{0}^{l} \mathbf{h}_{i}^{l}\right)
\label{eq:rgcn}
\end{equation}
where \(\mathcal{N}_{i}^{r}\) denotes the one-hop neighbors of \(i\)-th node that have edges of relation \(r\) and \(\mathbf{h}_{i}^{l} = \mathbf{H}^l[i,:] \in \mathbb{R}^{1*d^l}\) (\(d^l\) is the embedding dimension). \(c_{i, r} = |\mathcal{N}_{i}^{r}|\) is the normalized term, \(\sigma\) is the Sigmoid activation function, and \(\mathbf{W}_{*}^{l}\) are learnable weight matrices in the \(l\)-th layer.

After one-step neighborhood propagation via Eq. (\ref{eq:rgcn}), we obtain the contextual information in one-hop neighborhood. By stacking \(k\) layers of R-GCN, we can aggregate the information from \(k\)-hop neighborhood, where \(k\) is a hyperparameter that will be discussed in the experimental section. We denote the final output of \(k\)-layer R-GCN as \(\mathbf{H} \in \mathbb{R}^{N*d}\), where \(N\) and \(d\) are the number of nodes and the embedding dimension, respectively.

\subsection{Evidence-level Representation Readout}
The word-level node representations \(\mathbf{H}\) are processed using a readout module to generate evidence-level embeddings. Inspired by the work in the graph classification \cite{ying2018hierarchical}, we employ both max pooling and mean pooling over the words within each evidence to produce its representation.
\begin{equation}
    \mathbf{e}_m = \text{max}(\mathbf{H}_{i:j}) \ || \ \text{mean}(\mathbf{H}_{i:j})
\end{equation}
where \(\mathbf{H}_{i:j} \in \mathbb{R}^{(j-i+1)*d}\) denotes the segment of \(\mathbf{H}\) spanning rows $i$ to $j$, corresponding to nodes from the \(m\)-th evidence. The pooling strategies are applied along the first dimension so as to obtain the embedding of each evidence \(\mathbf{e}_m \in \mathbb{R}^{1*2d}\).    

In contrast to the graph classification context where pooling is performed over all graph nodes, this evidence-wise readout scheme proves beneficial for our task due to the varying significance of different evidence pieces.

\subsection{Attention-based Claim-evidence Interaction}
Evidence representations \(\{\mathbf{e}_1, \mathbf{e}_2, \ldots, \mathbf{e}_M\}\) have varying significance for claim verification. For example, the imperfect upstream evidence retrieval model may recall some redundant, claim-unrelated evidence. Such evidence should be ignored in the reasoning model. According to this observation, we introduce an attention-based claim-evidence interaction module. In detail, we compute the importance score \(\alpha_m\) for the \(m\)-th evidence regarding the claim based on an attention mechanism,
\begin{align}
\mathbf{c} &= \text{PLM}(c) \\
g_{m} &=\mathbf{W}_{a1}\left(\operatorname{ReLU}\left(\mathbf{W}_{a0}\left(\mathbf{c} \| \mathbf{e}_{m}\right)\right)\right) \label{equa:gm}\\
\alpha_{m} &=\operatorname{softmax}\left(g_{m}\right)=\frac{\exp \left(g_{m}\right)}{\sum_{i=1}^{M} \exp \left(g_{i}\right)} 
\end{align}
where PLM denotes a RoBERTa model encoding the claim into embeddings. PLMs in this module and node construction share the same weights for efficiency. We take the [CLS] representation as the claim embedding \(\mathbf{c} \in \mathbb{R}^{1*d}\). 
We then obtain the graph representation of the whole evidence set \(\mathbf{o}_g \in \mathbb{R}^{1*2d}\) via the attention-weighted summation of all evidence.
\begin{equation}
    \mathbf{o}_g =\sum_{i=1}^{M} \alpha_{i} \mathbf{e}_{i}
\end{equation}

\subsection{Fused Veracity Prediction}
 The graph construction method utilized here effectively captures the interaction between evidence, but it also weakens the integrity of the claim and evidence paragraphs. We found that relying solely on the graph representation may cause the model to overlook phrases such as negation words. Therefore, in addition to solely using the graph representation, we generate an assisted representation \(\mathbf{o}_t\) by feeding the linearized claim and evidence sequences directly to a PLM.  Then, the graph representation \(\mathbf{o}_g\) is concatenated with the assisted representation \(\mathbf{o}_t\) and fed into a multi-layer fully-connected network, followed by softmax normalization, to produce the final prediction \(\hat{\mathbf{p}} \in \mathbb{R}^{1*C}\), where \(C\) denotes the number of class.
\begin{align}
    \mathbf{o}_t &= \text{PLM}(\left[c, e_1,e_2,\cdots,e_M\right])\\
    \hat{\mathbf{p}} &= \operatorname{softmax}(\operatorname{MLP}(\mathbf{o}_g \| \mathbf{o}_t)) 
\end{align}

\subsection{Model Training}
 The cross-entropy objective is utilized to compute the veracity classification loss,
 \begin{equation}
Loss_c=-\sum_{c=1}^{C} \mathbf{y}_{c} \log \left(\hat{\mathbf{p}}_{c}\right)      
 \end{equation} 
where \(\mathbf{y}\) denotes the one-hot label vector (e.g., [1, 0, 0] represents that the ground truth is the first class). 

To counteract the effects of the upstream evidence retrieval model's imperfections, we utilize the advantageous attributes of attention mechanisms in the attention-based claim-evidence interaction module.
Specifically, we calculate a predicted evidence score \(s_m\) for each evidence \(e_m\) using the sigmoid function, and then compare it with the truth evidence label to generate a binary classification loss. The truth evidence label indicates whether an evidence is relevant for verifying the claim. The assisted loss is obtained by averaging all binary classification losses over the evidence set \(\mathcal{E}\):
\begin{align}
    s_m &= \operatorname{sigmoid}(g_m) \\
    Loss_e = \frac1M\sum_{i=1}^{M}&\left[t_m\log(s_m)+(1-t_m)\log(1-s_m) \right]
\end{align}
where \(g_m\) is a intermediate result in computing attention scores in Equation \ref{equa:gm}, \(t_m\) denotes the truth evidence label for evidence \(e_m\).

For training the whole model, we combined the loss item \(loss_c\) and \(loss_e\) with a hyperparameter \(\beta\), which will be discussed in the experimental section.
\begin{equation}
    Loss = Loss_c + \beta Loss_e
\end{equation}

\section{Experiment}
In this section, we conduct comprehensive experiments to answer the following research questions:
\begin{itemize}
    \item RQ1: How does the proposed method \themodel perform compared with existing baselines?
    \item RQ2: How is the word-level graph compared with the sentence-level graph and the token-level graph?
    \item RQ3: How does the model perform with different design of edges?
    \item RQ4: Which part of the model contributes most to the final result apart from the graph construction?
    \item RQ5: How does the model performance change with different values of hyper-parameters? 
    \item RQ6: How does the model perform with different retrieval model?
\end{itemize}

\subsection{Experimental Setups}

\subsubsection{Dataset}
Following prior research, we utilize the extensive FEVEROUS dataset in our experiments \cite{aly2021feverous}. The FEVEROUS task involves finding relevant evidence before claim verification. Each claim is manually annotated with labels: Supported, Refuted, or Not Enough Information, and paired with a corresponding golden evidence set. The dataset is divided into a training set of 71,291 claims, a development set of 7,890 claims, and a blind test set available on an online judging system\footnote{https://eval.ai/web/challenges/challenge-page/1091/overview}. The evidence sets include 38,941 with sentences only, 30,574 with tables only, and 25,395 with both sentences and tables.

\subsubsection{Metrics} 
Two metrics gauge the model's performance: the Feverous score and label accuracy. Label accuracy solely evaluates claim veracity classification accuracy, whereas the Feverous score assesses verdict prediction accuracy and correct retrieval of the golden evidence set. This score quantifies instances where the golden evidence set is successfully retrieved and the verdict is correctly predicted. The Feverous score is a comprehensive metric that assesses both the retrieval system and veracity reasoning model's performance.

\begin{table*}[t]
  \centering
  \resizebox{0.85\textwidth}{!}{
    \begin{tabular}{cccccc}
    \toprule
          & \multirow{2}[4]{*}{Model} & \multicolumn{2}{c}{Dev} & \multicolumn{2}{c}{Test} \\
\cmidrule{3-6}          &       & Feverous & Label Accuracy & Feverous & Label Accuracy \\
    \midrule
    \multirow{3}[2]{*}{Transformer-based} & $\text{RoBERTa-Pair}_{\text{mean}}$ & 0.3452 & 0.7117 & 0.3231 & 0.6074 \\
          & $\text{RoBERTa-Pair}_{\text{max}}$ & 0.3550 & 0.7207 & 0.3347 & 0.6190 \\
          & RoBERTa-Concat & 0.3549 & 0.7221 & 0.3344 & 0.6194 \\
          & DCUF \cite{hu2022dual} & 0.3577 & 0.7291 & 0.3397 & 0.6321 \\
    \midrule
    \multirow{3}[2]{*}{Graph-based} & \(\text{GEAR}\) \cite{zhou2019gear} & 0.2640 & 0.5859 & 0.2483 & 0.4964 \\
    & \(\text{KGAT}\) \cite{liu2020fine} & 0.3293 & 0.6844 & 0.3043 & 0.5797\\
          
          & \themodel & \textbf{0.3714} & \textbf{0.7352} & \textbf{0.3476} & \textbf{0.6329} \\
    \bottomrule
    \end{tabular}%
}
\caption{
  Comparison of models on Feverous task\\
  }
  \label{tab:main results}
\end{table*}%

\subsubsection{Baselines}
\begin{itemize}
    \item \(\textbf{RoBERTa-Pair}_{\text{mean/max}}\)
        Utilizes RoBERTa as backbone. Concatenates claim with each evidence separately to form sentences. Mean or max pooling is applied over embeddings of all evidence for final prediction.
    \item \textbf{RoBERTa-Concat}
        Concatenates claim with all evidence using [SEP] as separator. [CLS] token's representation is used for classification.
    \item \textbf{GEAR} \cite{zhou2019gear}
        Homogeneous coarse-grained graph-based method. Treats each claim-evidence pair as node in a fully-connected evidence graph. Graph convolutional network and evidence aggregator are used.
    \item \textbf{KGAT} \cite{liu2020fine}
        Kernel graph attention model. Similar graph construction as GEAR. Employs node and edge kernels for fine-grained evidence propagation.
    \item \textbf{DCUF} \cite{hu2022dual}
        Dual-channel approach. Converts evidence to sentence form and table form. Utilizes RoBERTa for sentence form, and TAPAS for table form. Integrates both channels for veracity prediction.
\end{itemize}

\subsubsection{Evidence Retrieval}
The primary objective of our paper is to present a model for veracity reasoning. For equitable comparison, we employ the evidence retrieval model from \cite{hu2022dual} for all tested models. We retrieve up to 150 relevant Wikipedia pages per claim using entity-matching and TF-IDF. The top 5 pages are selected based on SBERT\footnote{https://huggingface.co/cross-encoder/ms-marco-MiniLM-L-12-v2} and BM25 rankings. Tables within these pages are flattened, and up to 5 sentences and 3 tables are selected per claim using DrQA \cite{chen2017reading}. Relevant table cells are identified using a cell selector. Each claim's retrieved evidence set comprises up to 5 sentences and 25 table cells, as FEVEROUS task requirements.

\subsubsection{Implementation Details}
To boost our model's ability to extract relevant details from noisy evidence,  we augment each claim with two sets of evidence: the golden evidence set and the retrieved evidence set. 
% This augmentation doubles the training set size. 
Claims with the golden evidence set have all truth evidence labels set to positive. while for claims with the retrieved evidence set, only evidence shared with the golden evidence set receives a positive label. We consistently apply this augmentation to all baseline models. This approach differs from GEAR and KGAT's original technique for the FEVER task \cite{thorne2018fever}, which focuses on fact verification over unstructured text evidence.
 We use the Adam optimizer \cite{kingma2015adam} with learning rates of 1e-5 for language model parameters and 1e-3 for others, employing a linear scheduler with a 20\% warm-up rate. RoBERTa-Large serves as the PLM, with window size $w$ and R-GCN layer count $k$ set to 2. All experiments run on a server with an AMD EPYC 7742 (256) @ 2.250GHz CPU and one NVIDIA A100 GPU.

\subsection{Overall Performance (RQ1)}
We compare our proposed \themodel against various baselines, spanning transformer-based and graph-based models. In Table \ref{tab:main results}, \themodel consistently outperforms all strong baselines across metrics and datasets, highlighting its superiority. Key observations from these results are as follows:
\begin{itemize}
    \item  Among RoBERTa-based models, \(\text{RoBERTa-Pair}_{\text{mean}}\) lags, while the other two exhibit similar performance levels. This may stem from RoBERTa-Pair's limited ability to capture diverse relationships by processing evidence pieces separately. RoBERTa-Concat processes evidence together, but struggles to distinguish between golden and noisy evidence, affecting its performance. Thus, direct PLM use falls short, emphasizing the need for task-specific design.
    \item DCUF, the top-performing transformer-based method, incorporates RoBERTa-Concat alongside TAPAS, contributing to its superior performance.
    \item Graph-based models GEAR and KGAT, stemming from FEVER task, exhibit suboptimal performance due to task and data differences. Comparing them with \(\text{\themodel}_{\text{graph}}\) (see Table \ref{tab:ablation}), which is a purely graph-based model, \(\text{\themodel}_{\text{graph}}\) outperforms substantially, affirming the efficacy of our hybrid fact verification design.
\end{itemize}

\subsection{Ablation Study and Model Variants}
\textbf{Comparison of the Graph Granularity (RQ2) \ }
\label{sec:granularity}
\begin{figure}[t]
\setlength{\abovecaptionskip}{0pt}  % reduce gap
\setlength{\belowcaptionskip}{0pt}  % reduce gap
   \begin{center}
  \vspace{-0mm}
   \includegraphics[width=1\columnwidth]{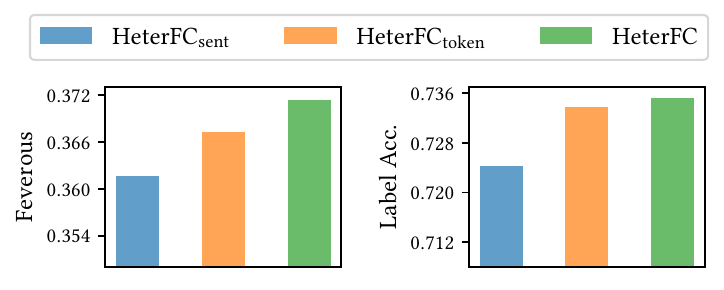}
   \end{center}
   \vspace{0cm}
   \caption{The performance comparison among models with different graph granularity. $\text{\themodel}_{\text{sent}}$ represents the variant with sentence-level graph, while $\text{\themodel}_{\text{token}}$ represents the variant with token-level graph. } 
   \label{fig:grain}
\end{figure}
\begin{figure}[t]
\setlength{\abovecaptionskip}{0pt}  % reduce gap
\setlength{\belowcaptionskip}{0pt}  % reduce gap
   \begin{center}
  \vspace{0mm}
   \includegraphics[width=1\columnwidth]{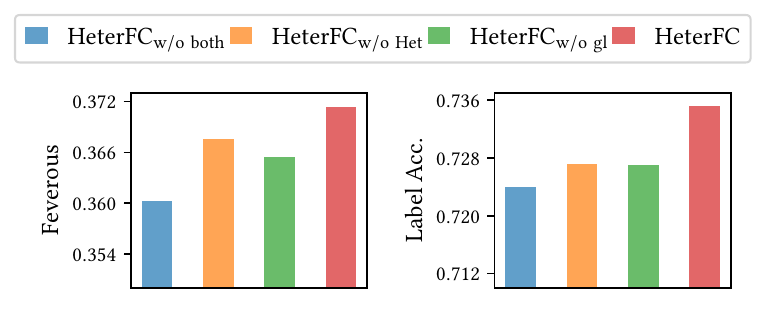}
   \end{center}
   \vspace{0cm}
   \caption{The performance comparison between the proposed model \themodel and its three variants: $\text{\themodel}_{\text{w/o gl}}$ ignores global and local connection designs. $\text{\themodel}_{\text{w/o Het}}$ ignores heterogeneous relations. $\text{\themodel}_{\text{w/o both}}$ removes all special designs for a fully-connected homogeneous graph. } 
   \label{fig:edge}
\end{figure}
To validate our word-level graph design, we experiment with \themodel against its sentence-level graph variant ($\textbf{\themodel}_{\text{sent}}$) and token-level graph variant ($\textbf{\themodel}_{\text{token}}$). For sentence-level graphs, each claim-paired sentence (tables linearized into sentences) constitutes a node. Nodes from the same claim interconnect to form a fully-connected evidence graph, initialized with PLM [CLS] token embeddings. For token-level graphs, we omit subword mean pooling, resulting in nodes representing subwords.

From Figure \ref{fig:grain}, \themodel surpasses both model variants across metrics. $\text{\themodel}_{\text{sent}}$ significantly lags behind \themodel and $\text{\themodel}_{\text{token}}$. This gap suggests that $\text{\themodel}_{\text{sent}}$'s coarser granularity struggles to capture nuanced semantic relationships among evidence pieces.
Comparing \themodel with $\text{\themodel}_{\text{token}}$, \themodel maintains better Feverous score and slightly edges in label accuracy. This suggests that the word-level graph in \themodel enhances performance due to more precise inter-evidence connections than the token-level graph. In token-level graphs, shared subwords may lead to extraneous inter-evidence links among unrelated words (e.g., ``interesting'' and ``thing'' sharing ``ing''). This results in noisy connections due to non-keyword shared subwords, undermining overall performance.

\bigskip
\noindent\textbf{Comparison of Different Edge Construction Strategies (RQ3) \ }
In \themodel, unique edge constructions include heterogeneous relations, inter-evidence global connections, and intra-evidence local connections. This section ablates these strategies through model variants to assess their impact. For $\textbf{\themodel}_{\text{w/o Het}}$, homogeneous edges replace heterogeneous relations, making all edges identical. $\textbf{\themodel}_{\text{w/o gl}}$ disregards specific local and global connection designs, yielding a fully-connected graph while retaining heterogeneous relations. $\textbf{\themodel}_{\text{w/o both}}$ discards all special designs, leading to a fully-connected and homogeneous graph.

From Figure \ref{fig:edge}, \themodel outperforms its variants, with $\text{\themodel}_{\text{w/o both}}$ exhibiting the weakest performance. Furthermore, $\text{\themodel}_{\text{w/o gl}}$ and $\text{\themodel}_{\text{w/o Het}}$ show marked performance drops, emphasizing the effectiveness of both heterogeneous edges and local/global connection designs.

\begin{table}[tbp]
  \centering
  \resizebox{0.76\columnwidth}{!}{
    \begin{tabular}{ccc}
    \toprule
    Model & Feverous & Label Accuracy \\
    \midrule
     \(\text{\themodel}\) &    \textbf{0.3714}   &  \textbf{0.7352}\\
    \midrule
     \(\text{\themodel}_{\text{mean}}\) &    0.3613   & 0.7238 \\
     \(\text{\themodel}_{\text{max}}\) &    0.3620   &  0.7226\\ 
     \(\text{\themodel}_{\text{graph}}\) &   0.3640    & 0.7243 \\
     \(\text{\themodel}_{\text{single}}\) &    0.3641   &  0.7257\\
    \bottomrule
    \end{tabular}%
    }
    \caption{Comparison between \themodel and its variants: \(\text{\themodel}_{\text{mean/max}}\) substitutes attention-based claim-evidence interaction with mean/max pooling. \(\text{\themodel}_{\text{graph}}\) uses only graph representation for a purely graph-based model. \(\text{\themodel}_{\text{single}}\) ignores the assisted loss. }
  \label{tab:ablation}%
\end{table}%

\bigskip
\noindent\textbf{Investigating Attention Module, Veracity Prediction and Losses (RQ4) \ }
In addition to the discussed graph construction, we further evaluate the Attention-based Claim-evidence Interaction model, Fused Veracity Prediction, and assisted loss designs through several \themodel variants:
\begin{itemize}
\item \(\textbf{\themodel}_{\text{mean/max}}\): Substitutes the original Attention-based Claim-evidence Interaction with a mean/max pooling layer for evidence representation aggregation. No assisted loss is computed due to the absence of predicted evidence scores.
\item \(\textbf{\themodel}_{\text{graph}}\): Excludes \(\text{o}_t\) in Fused Veracity Prediction, relying solely on graph representation \(\text{o}_g\) for a purely graph-based model.
\item \(\textbf{\themodel}_{\text{single}}\): Using only the veracity classification loss \(Loss_c\), ignoring the assisted loss \(Loss_e\).
\end{itemize}
Table \ref{tab:ablation} displays development set results for these variants. Notably, \themodel surpasses all variants.  \(\text{\themodel}_{\text{mean/max}}\) perform worse than \(\text{\themodel}_\text{{single}}\). These three variants lack assisted loss during training, highlighting the importance of the Attention-based Claim-evidence Interaction model. Comparing \(\text{\themodel}_\text{{single}}\) and \(\text{\themodel}_\text{{graph}}\), assisted loss and Fused Veracity Prediction contribute similarly to outcomes. The pivotal role of the Attention-based Claim-evidence Interaction module is evident.

\begin{figure}[t]
\setlength{\abovecaptionskip}{0pt}  % reduce gap
\setlength{\belowcaptionskip}{0pt}  % reduce gap
   \begin{center}
  % \vspace{-4mm}
   \includegraphics[width=1\columnwidth]{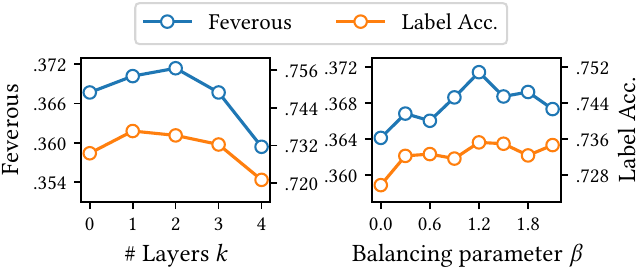}
   \end{center}
   \vspace{0cm}
   \caption{The influence of hyper-parameters on \themodel's performance on the development set.} 
   \label{fig:hyperparam}
\end{figure}

\subsection{Hyperparameter Sensitivity Analysis (RQ5)}
We study the impact of two key hyperparameters: the number of R-GCN layers \(k\) determining information aggregation range and  the parameter $\beta$ for balancing \(Loss_e\) and \(Loss_c\).

\subsubsection{Number of layers \(k\).} In the left part of Figure \ref{fig:hyperparam}, increasing GNN layers enhances metrics, but beyond $k=2$, metrics drop due to over-smoothing. We select $k=2$ for \themodel.
 
\subsubsection{Balancing parameter $\beta$.}The right part of Figure \ref{fig:hyperparam} shows Feverous score peaks at 
$\beta=1.2$ before a slight decline, while label accuracy increases steadily with larger $\beta$, stabilizing eventually. We select $\beta=1.2$ as optimal for \themodel.

\subsection{Evaluating the Robustness of \themodel (RQ6)}

\begin{table}[tbp]
  \centering
  
  \resizebox{0.782\columnwidth}{!}{
    \begin{tabular}{ccc}
    \toprule
           Model & Feverous & Label Acc. \\
    \midrule
            $\text{RoBERTa-Pair}_{\text{mean}}$     & 0.2272     & 0.6689 \\
           $\text{RoBERTa-Pair}_{\text{max}}$   &  0.2379    & 0.6829 \\
           RoBERTa-Concat    & 0.2352     & 0.6627 \\
           DCUF     & 0.2380     & 0.6895 \\
    % \midrule 
           GEAR     & 0.1783     & 0.5786 \\
           KGAT     & 0.2110     & 0.6117 \\
           \themodel     & \textbf{0.2427}     & \textbf{0.6977} \\
    \bottomrule
    \end{tabular}%
    }
    \caption{Comparison of model performances on the development set under different retrieval method. }
  \label{tab:other_retrieve}%
\end{table}%

We extend our analysis beyond the retrieval method by \cite{hu2022dual} to include a simpler approach by \cite{aly2021feverous}. Despite less effective retrieval, it simulates scenarios with limited, noisy evidence. Results in Table \ref{tab:other_retrieve} align with those in Table \ref{tab:main results}, showing that \themodel maintains superior performance over baseline models, underscoring its robustness even with less optimal retrieval techniques.

% The preceding analyses rely on the retrieval method by \cite{hu2022dual}, offering limited insights. To broaden our understanding of \themodel's performance, we extended assessments to a simpler retrieval method by \cite{aly2021feverous}. This approach yields less effective retrieval but simulates scenarios with limited, noisy evidence.
% Results on the development set are in Table \ref{tab:other_retrieve}. Comparing with Table \ref{tab:main results}, we observe similarities. \themodel continues to outperform baseline models, reaffirming its robustness even with suboptimal retrieval techniques.

\section{Conclusion}

In this paper, we introduce \themodel, a novel word-level heterogeneous-graph-based model for fact checking that effectively combines unstructured and structured information. Our model employs a carefully designed graph structure with word-level nodes and diverse edge types. We integrate a heterogeneous information propagation module with attention-based claim-evidence interaction to capture the semantic relationships between claims and evidence. Additionally, we introduce an assisted loss based on attention scores to differentiate valid and invalid evidence. Extensive experiments confirm the superiority of \themodel over diverse baseline models.

\section{Acknowledgements}
This work is jointly sponsored by National Natural Science Foundation of China (U19B2038, 62372454, 62206291, 62236010).

\bibliography{aaai24}

\end{document}